\documentclass[runningheads]{llncs}
\usepackage{booktabs}
\usepackage{multirow}
\usepackage[table]{xcolor}
\usepackage[T1]{fontenc}
\usepackage{tikz}
\usepackage[nocompress]{cite}
\usepackage[labelfont=bf]{caption}
\usepackage{subcaption}
\usepackage{url}
\usepackage{graphicx}
\usepackage[colorlinks=true]{hyperref}
\usepackage{color}
\hypersetup{
    linkcolor=blue,  
    citecolor=blue,  
    urlcolor=blue    
}

\begin{document}

\newcommand{\benchname}{PorTEXTO}
\newcommand*{\diogo}[1]{{\textcolor{red}{[DS: #1]}}}
\title{\benchname: A European \underline{Por}tuguese Benchmark for Visual \underline{T}ext \underline{Ext}rac{t}i\underline{o}n}
%
\titlerunning{\benchname}

 \author{João Cardeira\inst{1,2} \and
 Diogo Glória-Silva\inst{1,2} \and
 Manuel Letras da Luz\inst{1,2} \and
 Rafael Ferreira\inst{1,2} \and
 Diogo Tavares\inst{1,2} \and
 David Semedo\inst{1,2} \and
João Magalhães\inst{1,2}}
 \authorrunning{J. Cardeira et al.}
 \institute{NOVA School of Science and Technology \and
 NOVA LINCS \\
 \email{jaca.pereira@campus.fct.unl.pt}
}
\maketitle              
\setcounter{footnote}{0}
\begin{abstract}
European Portuguese (pt-PT) is largely absent from Optical Character Recognition (OCR) benchmarks, which skew toward high-resource languages. The few benchmarks that cover pt-PT focus on historical artifacts and literature. This work addresses modern OCR applications, introducing \benchname, the first benchmark for contemporary and culturally relevant pt-PT visual text extraction. To ascertain quality, we employ an annotation pipeline combining transcriptions from a frontier LVLM with exhaustive review by native speakers. We observe a sharp performance drop from synthetic to real world samples in most models, and find that, currently, specialized multilingual data is a better driver for pt-PT performance than model size or resolution budget, motivating the release of open pt-PT OCR resources.\footnote{Code is available at \url{https://github.com/AMALIA-LLM/amalia-vl-eval} and datasets at \url{https://huggingface.co/datasets/amalia-llm/PorTEXTO}.} 
\end{abstract}

\section{Introduction}

\begin{figure}[t]
    \centering
    \includegraphics[width=\linewidth]{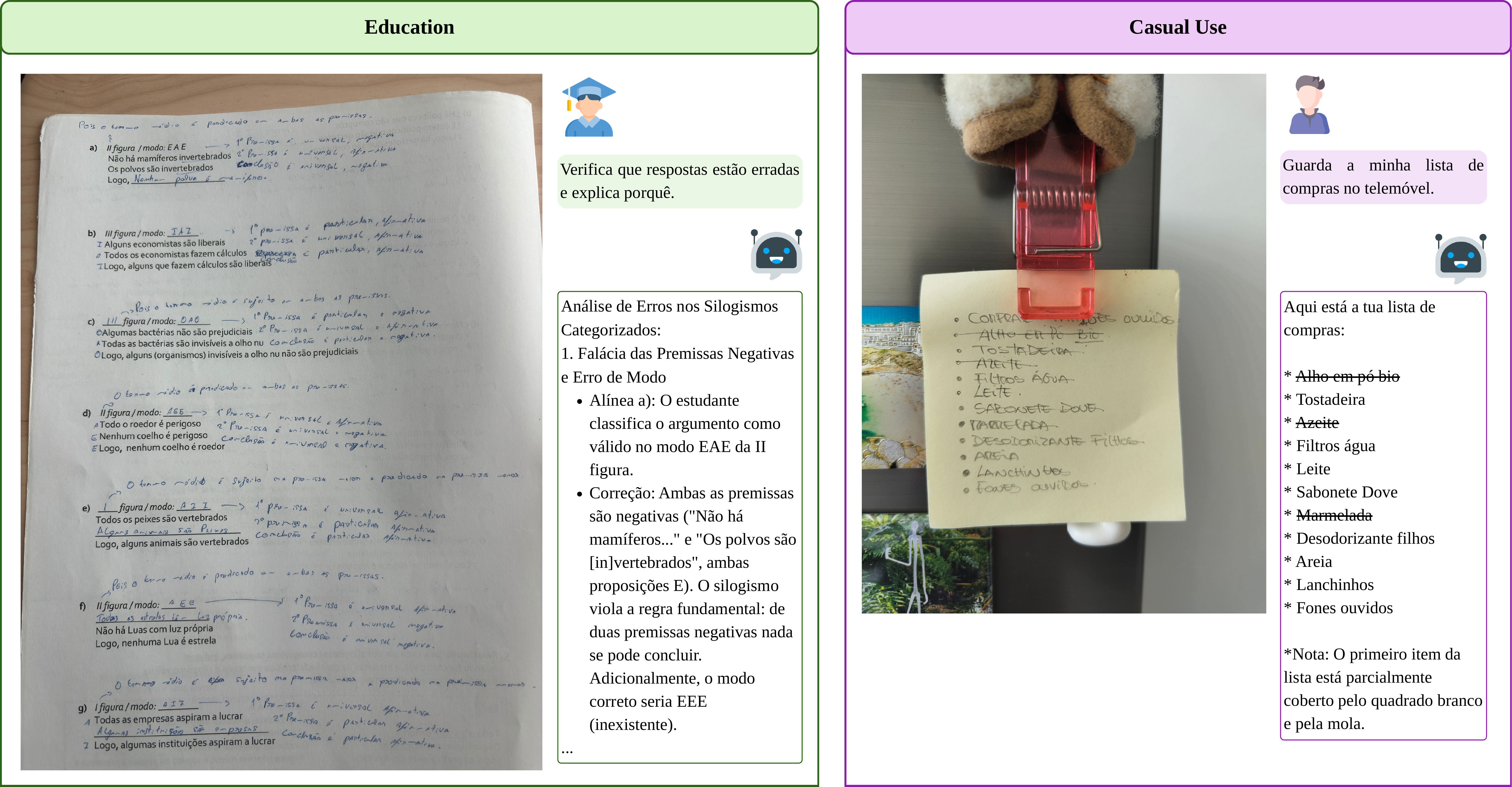}
    \caption{\textbf{Modern OCR applications.} Contemporary OCR applications range from professional settings, such as automatic document processing, to casual and educational purposes, often involving user dialogue with an AI chat.}
    \label{fig:modern_applications}
\end{figure}

European Portuguese (pt-PT) is  an important language, with native speakers in Portugal and across the world, yet remains significantly underexplored in Optical character recognition (OCR) benchmarks~\cite{ocrbench, docvqa, infovqa, omnidocbench,textvqa}, which are heavily skewed toward high-resource languages such as English (EN) and Chinese (ZH). Benchmarks that target pt-PT OCR~\cite{ester, porto, iforal, mazaafard} focus mostly on historical artifacts, spanning centuries during which Portuguese orthography changed significantly. Additionally, these datasets lack data representing modern OCR use cases (Figure~\ref{fig:modern_applications}), such as understanding text in casual photos or screenshots; automated contract, invoice and form processing for business applications; and exam and test processing for educational purposes. 
Therefore, the ability to visually recognize pt-PT text in modern dialect and contemporary OCR use cases remains unexplored.

\begin{figure}[t]
    \centering
    \includegraphics[width=\linewidth]{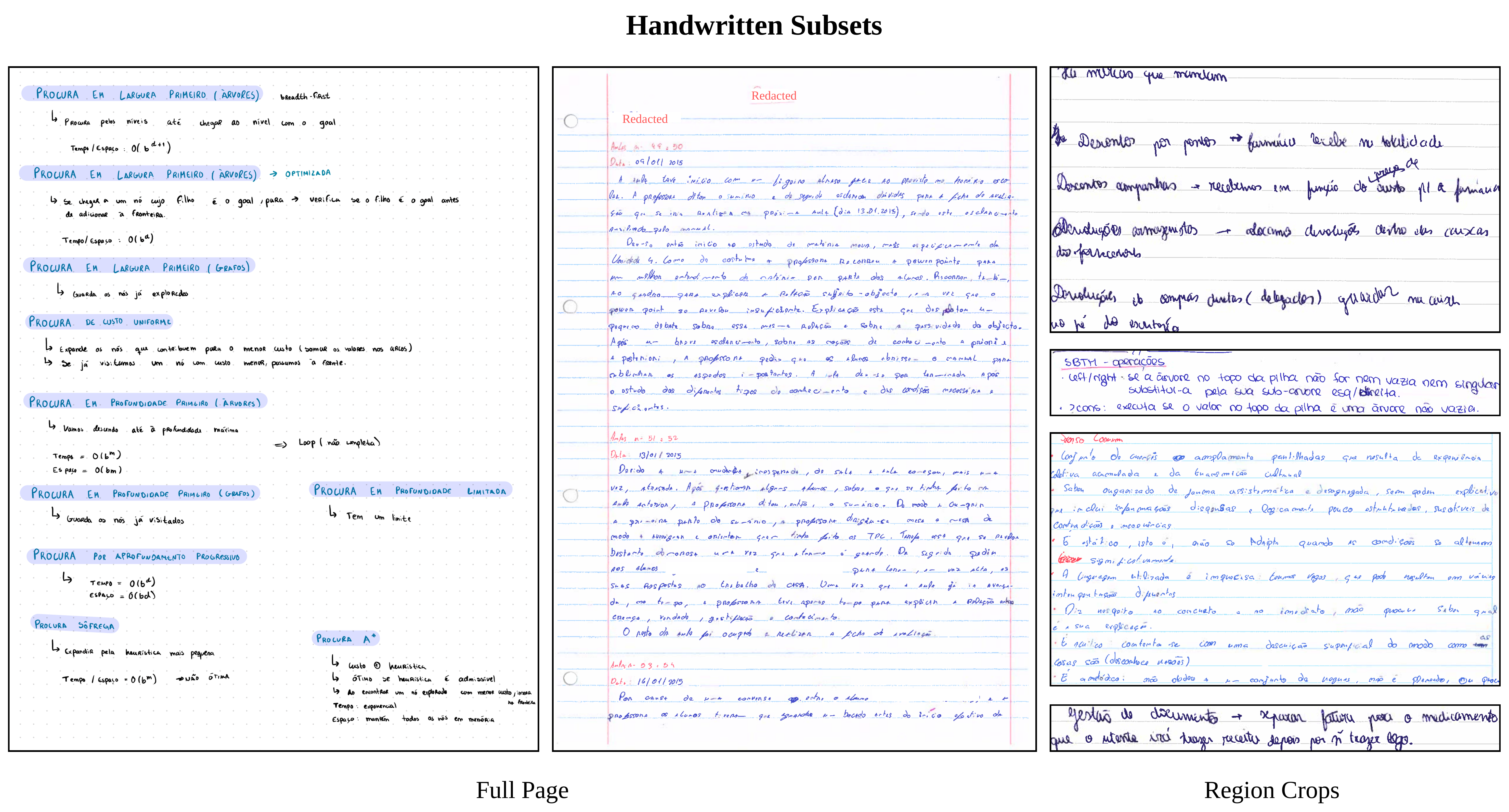}
    \caption{\textbf{Handwritten Subsets.} Representative samples from Full Page and Region Handwritten subsets of \benchname. The word \textit{redacted} is used purely for illustration purposes, signaling regions where human reviewers redacted poor text. These samples expose models to authentic handwriting, idiosyncratic pt-PT abbreviations, mixed prose, and everyday orthography.}
    \label{fig:handwritten_samples}
\end{figure}

\begin{figure}[t]
    \centering
    \includegraphics[width=\linewidth]{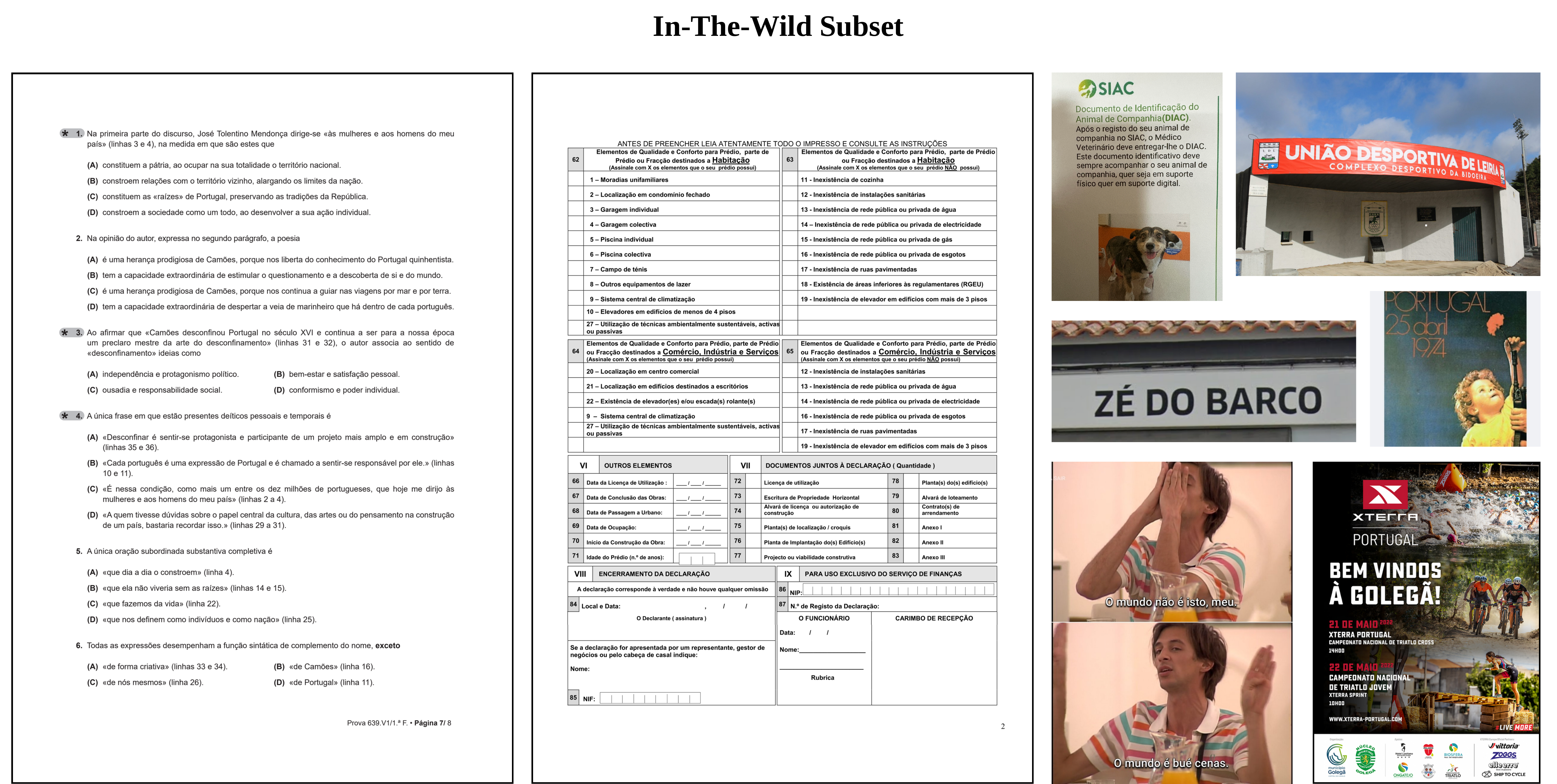}
    \caption{\textbf{In-The-Wild Subset.} Representative samples from In-The-Wild subset of \benchname. These samples expose models to important real world and culturally meaningful concepts. It includes a wide variety of domains, such as school exams, common invoices, government forms, and also a variety of samples of casual nature, such as photographs of coffee shops and town signs, popular sports clubs, TV shows and memes, and more. This subset also presents a challenging range of resolutions and aspect ratios.}
    \label{fig:in_the_wild_samples}
\end{figure}

\begin{figure}[t]
    \centering
    \includegraphics[width=\linewidth]{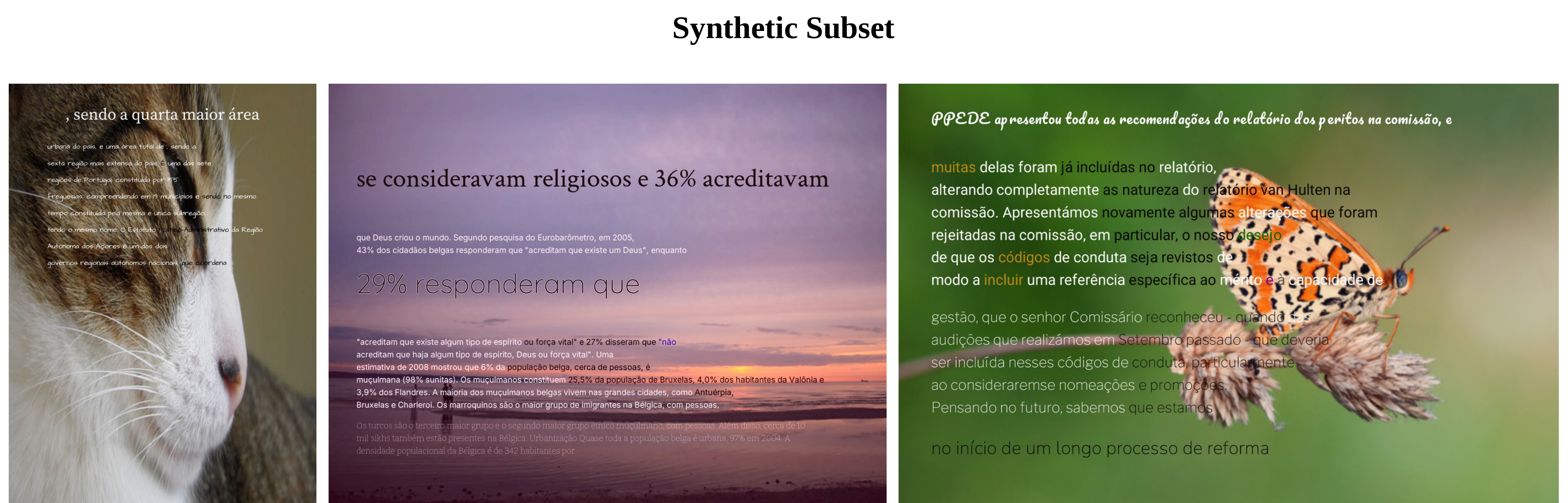}
    \caption{\textbf{Synthetic Subset.} Representative samples from the Synthetic subset of \benchname. These samples expose models to pure pt-PT text in common document layouts, rendered on top of general purpose images. Represents data more likely to be aligned with training data distribution of most models.}
    \label{fig:synthetic_samples}
\end{figure}

To address this gap, we introduce \textbf{\benchname}, a benchmark for pt-PT text extraction that explicitly targets contemporary usage across a broad range of culturally relevant use cases: (i) handwritten pt-PT text collected from student notes and personal writings, (ii) in-the-wild photos, screenshots and documents across common pt-PT domains, and (iii) synthetic images produced by a text rendering pipeline that composites pt-PT text over varied backgrounds.  We describe our annotation pipeline, where image text transcripts are annotated using a frontier LVLM and exhaustively reviewed and corrected by Portuguese annotators, ensuring quality. \textbf{\benchname}~exposes a significant gap between pt-PT OCR model performance in synthetic and real world samples, and the advantages of multilingual focus, motivating the creation of publicly available pt-PT-centric OCR resources. 

Our main contributions are the following: 1) we introduce \textbf{\benchname}, the first pt-PT image text extraction dataset covering contemporary language usage across diverse culturally relevant domains and backgrounds; 2) we describe an annotation pipeline that combines LVLM annotation with human reviews by native speakers, designed to ensure annotation quality; 3) we conduct the first systematic assessment of LVLM visual text extraction performance on modern pt-PT text.

\section{Related Work}

\vspace{-2mm}\subsubsection*{Multilingual Image Text Extraction.}

Despite the wide range of image and document-based text extraction benchmarks for high-resource languages such as English and Chinese~\cite{textvqa, okvqa, ocrbench, omnidocbench, docvqa, infovqa}, low-resource languages remain significantly underexplored. MTVQA~\cite{mtvqa} provides question--answer pairs that integrate visual reasoning over text-centric images in multiple languages, but lacks comprehensive OCR annotations. %
CC-OCR~\cite{cc-ocr} evaluates the performance of LVLMs in literacy and includes a multilingual OCR subset spanning two East Asian languages, six Latin-script languages, one Cyrillic-script language and one Arabic-script language. Its Portuguese subset, however, is short and dominated by pt-BR vocabulary and Brazilian cultural references. 
BRESSAY~\cite{bressay} introduces a dataset for handwritten text recognition drawn from Brazilian student academic essays, leaving the pt-PT variety entirely unaddressed.

\vspace{-2mm}\subsubsection*{European Portuguese Image Text Extraction.}
EPISA~\cite{episa} provides 708 images with manually transcribed documents (i.e., letters, reports, theater play covers) from the 20th century. ESTER-Pt~\cite{ester} proposes an evaluation suite for text recognition in pt-PT, containing synthetic image and text-based documents derived from the Portuguese Wikipedia and real scanned books written between 1800 and 1910. PORTO~\cite{porto} further extends this dataset with historical pt-PT documents from the 17th to the 20th centuries, yielding 3,782 image--transcription pairs. Following a similar approach, iFORAL~\cite{iforal} proposes a dataset of transcribed pt-PT medieval documents, focused on transcribing municipal charters from the Middle Ages. 
Portuguese-OCR~\cite{mazaafard} synthetically renders single-sentence text passages from classic pt-PT literature into images, using different backgrounds and fonts. These resources cover historical artifacts and literary excerpts under limited visual variability. Thus, the behavior of modern models on contemporary pt-PT visual text remains essentially unaddressed.

Unlike previous works, \benchname~is a pt-PT--centric benchmark that evaluates text extraction from contemporary and culturally-relevant images and documents across a wide variety of scenarios and difficulty levels.

\section{\benchname~Methodology} 

We designed our data collection and curation methodology around the goal of representing modern OCR use cases within the European Portuguese culture and dialect. We relied on voluntary effort from contributors for data collection, and on seventeen native European Portuguese speakers for data curation.

\subsection{Collecting Contemporary and Culturally Relevant Data}
\label{sec:data_collection}

\benchname~contains samples from three complementary sources covering modern visual and linguistic contexts in which pt-PT text appears in practice. 

\vspace{-2mm}\vspace{-2mm}\subsubsection{Handwritten Notes.} We collected personal handwriting, including day-to-day notes, as well as high-school and college-level notes. These cover general daily topics (e.g., grocery shopping lists) and more advanced topics from engineering, social sciences, among others.  
This subset represents typical use cases of pt-PT speakers (e.g., transcribing their personal notes to a digital format), exposing models to authentic handwriting, idiosyncratic abbreviations, mixed prose, and the everyday orthography of contemporary pt-PT. Each contribution was digitally scanned into a multi-page A4 PDF at 200 dpi, which was then rasterized into one high-resolution PNG per page that served as raw input to the annotation pipeline. Representative samples can be viewed in Figure~\ref{fig:handwritten_samples}.

\vspace{-2mm}\subsubsection{In-the-wild Images.}
The In-The-Wild subset includes samples more aligned with casual applications, such as photographs of town signs, building names, festivity posters, restaurant menus, personal notes, screenshots of popular memes; and also samples representative of business or educational applications, such as documents, exams, forms, and invoices. In addition to these scenarios, which are largely absent from existing pt-PT benchmarks, this subset also introduces different visual factors such as variable noise, lighting, perspectives, and fonts. Representative samples can be viewed in Figure~\ref{fig:in_the_wild_samples}.

\begin{figure}[t]
    \centering
    \includegraphics[width=\columnwidth]{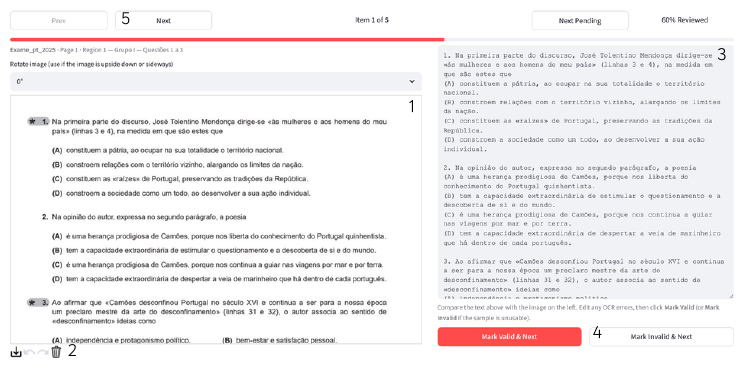}
    \caption{The platform used for annotation. Annotators can visualize the image (1); redact the image by drawing a square using the mouse pointer, and can reverse redactions (2); verify and correct transcriptions (3); validate or invalidate samples (4); and iterate to the next or the previous sample, as well as visualize progress (5).}
    \label{fig:annotation-platforms}
  \end{figure}

\vspace{-2mm}\subsubsection{Synthetic Images.}
The Synthetic subset contains pt-PT text rendered in background images. To guarantee quality and pt-PT focus, we source text from a Portuguese Wikipedia corpus~\cite{wikimedia2023wikipedia} and the pt-PT subset of transcripts from the European Parliament~\cite{koehn2005europarl}. We used a diverse collection of open-source, Portuguese-compatible fonts. The text was rendered using SynthDoG~\cite{synthdog}, a popular document generation pipeline, with structurally varied document layouts, including tables of contents, presentation slides, and multi-section document pages, each imposing distinct spatial arrangements of headings, body paragraphs, indented entries, or bullet points. These layouts were placed over background images sampled from the Open Images~\cite{openimages} dataset. To avoid rendering text over background with identical color, we measure contrast between font color and background pixels and dynamically adapt color at word level. Representative samples can be viewed in Figure~\ref{fig:synthetic_samples}.

\subsection{Curation of Reference Transcriptions}
\label{sec:data_annotation}
To enforce data quality, collected data was filtered by an ensemble of Portuguese human reviewers in two stages of the pipeline. The interface provided to the annotators can be seen in Figure~\ref{fig:annotation-platforms}.

\vspace{-2mm}\subsubsection*{Sample Reviewing and Cleaning.} Annotators inspected images from the Handwritten and In-The-Wild subsets, invalidating samples where the text was mostly blurry or noisy, and could not be read by humans, or samples that are almost entirely mathematical notation, since the latter falls outside the scope of \benchname's pt-PT text-extraction focus. Reviewers were allowed to recover samples by redacting only portions of the text that they could not reliably comprehend.

\vspace{-2mm}\subsubsection*{Automatic OCR Transcripts.}
The handwritten notes and in-the-wild images were annotated by a frontier LVLM, Gemini~3.1 Pro Preview~\cite{gemini}. We chose it over standard OCR methods due to its strong OCR performance and proven solid coverage of pt-PT~\cite{euroeval}. The model was prompted to faithfully transcribe each image into plain text, preserving spatial layout, and to encode any remaining mathematical expressions in LaTeX, for samples where their presence is minimal. Synthetic images were exempt from annotation since transcripts are correct by their nature.

\vspace{-2mm}\subsubsection{Handwritten Data Augmentation.}
To increase the granularity of the Handwritten subset, each scanned page was augmented by splitting it into several sub-regions of text pertaining to unique subtopics. These sub-regions were obtained by prompting the same LVLM to return, for each page, an ordered list of bounding boxes corresponding to its main logical content blocks, in natural reading order. The page was then cropped according to each bounding box, yielding region-level samples that were subsequently filtered and transcribed as aforementioned, effectively branching Handwritten into two subsets: Handwritten (Region) and Handwritten (Full Page) (see Figure~\ref{fig:handwritten_samples}).

\begin{figure}[t]
    \centering
    \includegraphics[width=\linewidth]{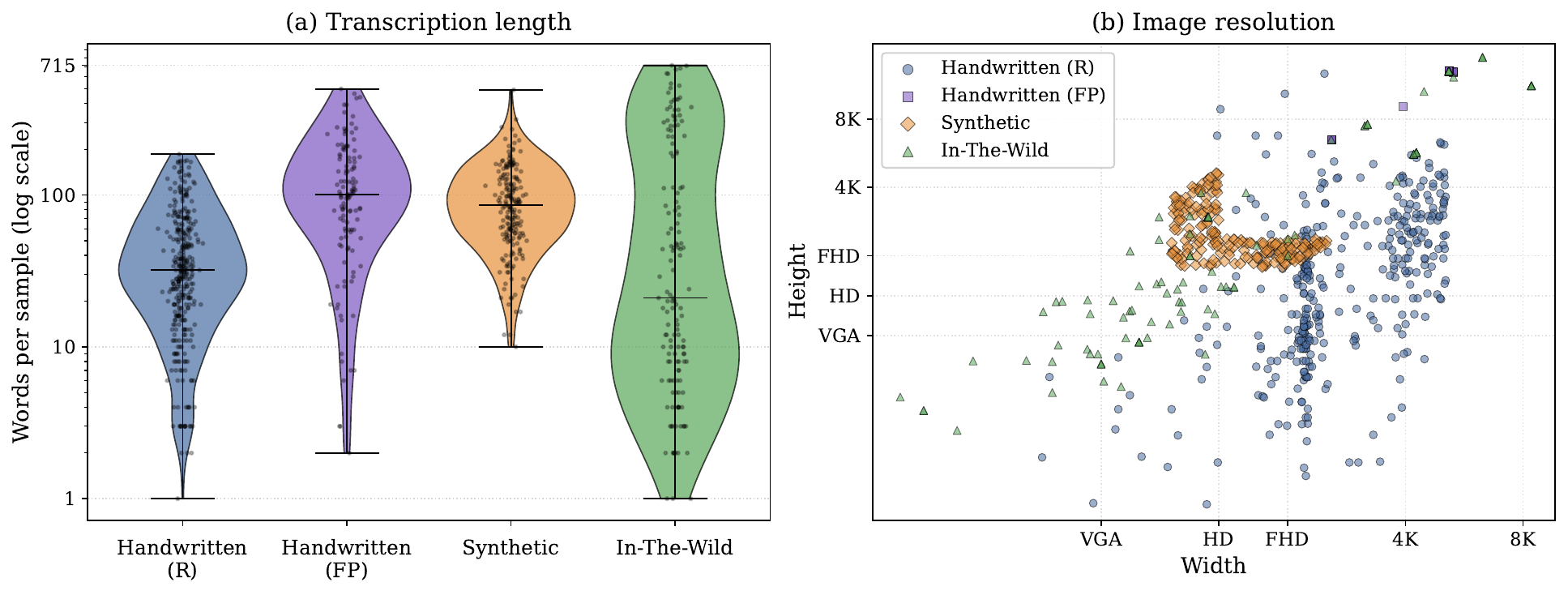}
    \caption{Distributions of transcription length and source image resolution. The wide spread along both dimensions reflects the diversity of textual content and visual conditions covered by the benchmark.}
    \label{fig:stats}
\end{figure}

\vspace{-2mm}\subsubsection*{Post-OCR verification.} Human reviewers performed a final correctness check on all transcribed samples, where each transcription was either validated, corrected, or marked as invalid and discarded. Samples were randomly distributed between annotators, increasing the likelihood that each sample was seen by a different reviewer from the first round, adding further guarantee that out-of-scope and poor samples are discarded.

\subsection{Curated Dataset Analysis}
\label{sec:dataset_analysis}
From a total of 873 collected samples, reviewers discarded 236 ($\approx27\%$) and redacted 175 ($\approx20\%$) samples. In total, 637 samples passed data cleaning. During Post-OCR verification, reviewers discarded 20 ($\approx3\%$) and corrected 135 ($\approx21\%$) samples. Post-curation, \benchname~comprises a total of 817 samples drawn from the three sources described above, with 121 handwritten pages and 351 region-level crops, 145 in-the-wild images, and 200 synthetic images.

Figure~\ref{fig:stats} summarizes the main statistics of the resulting benchmark, including transcription length and image-resolution distribution. The Handwritten subsets cover a large spectrum of transcriptions lengths, from short personal reminders and short cooking recipes to highly structured college and high-school level study notes. While the Handwritten (Region) subset covers a wide range of resolutions, reflecting the variable size of cropped topic blocks, the Handwritten (Full Page) subset clusters around three resolutions corresponding to scanned A4 pages. Images from the In-The-Wild subset also exhibit a high variety of transcription length and resolutions, resulting from the diversity of the samples. Photographs taken by volunteers have more squared resolutions typical from phones, and may contain fewer words; whereas an exam has a document-like resolution and a significantly higher amount of words. \benchname~has a total of 68,409 words, with a mean of 83.7 words per sample. Overall, the amplitude of scenarios, transcriptions, and resolutions reflects \benchname's diversity and challenges.

\section{Evaluation Setup and Metrics}

We adopted the \texttt{lmms-eval}~\cite{lmms_eval} platform, a unified evaluation framework that allows fair and reproducible evaluation. We selected a suite of models across three categories: OCR-specific models, First Tier LVLMs (8B to 12B) and Second Tier LVLMs (27B to 38B). We kept the default model configurations but set temperature to 0 for deterministic output. Thinking was disabled for unified models.

\begin{table}[t]
    \centering
    \scriptsize
    \setlength{\tabcolsep}{3pt}
    \renewcommand{\arraystretch}{1.2}      
    \caption{Image processing strategy and pixel budget range for evaluated models. \emph{Tiling}: actual spatial extent depends on aspect ratio; the budget is spread across fixed-size tiles. $^\dagger$Configurable; default shown. $^\ddagger$Longest-edge constraint (1540\,px); maximum resolution only achieved in square inputs.}
    \label{tab:model_properties}
    \begin{tabular}{lccc}
    \toprule
    \textbf{Model} & \textbf{Proc.} & \textbf{Min Px} & \textbf{Max Px} \\
    \midrule
    DeepSeek-OCR-2-3B~\cite{deepseek_ocr} & Tiling & 0.6M & 3.5M* \\
    Nemotron-OCR-V2~\cite{nemotron_ocr} & Fixed & 1.0M & 1.0M \\
    Ministral3-Instruct-8B~\cite{ministral3} & Adapt. & -- & 2.4M$^\ddagger$ \\
    PerceptionLM-8B~\cite{perceptionlm} & Tiling & 0.2M & 7.2M* \\
    GLM-4.6V-Flash-10B~\cite{glm} & Adapt. & 12.5K & 9.6M \\
    Molmo2-O-7B~\cite{molmo2} & Tiling & 0.1M & 1.1M* \\
    Molmo2-8B~\cite{molmo2} & Tiling & 0.1M & 1.1M* \\
    InternVL3.5-8B~\cite{internvl} & Tiling & 0.2M & 2.4M* \\
    AMALIA-VL~\cite{amalia-vl} & Tiling & 0.1M & 2.4M* \\
    AMALIA-VL-DPO~\cite{amalia-vl} & Tiling & 0.1M & 2.4M* \\
    Bee-RL-8B~\cite{beerl} & Tiling & 0.1M & 5.3M* \\
    Nemotron-Nano-12B-VL-V2~\cite{nemotron_vl} & Tiling & 0.3M & 3.1M* \\
    LLaVA-OneVision-1.5-8B~\cite{llavaonevision15} & Adapt. & 3.1K & 3.2M \\
    Gemma-4-E4B-it~\cite{gemma4} & Adapt. & 2.3K & 0.6M$^\dagger$ \\
    Qwen3.5-9B~\cite{qwen35} & Adapt. & 65.5K & 16.8M \\
    InternVL3.5-38B~\cite{internvl} & Tiling & 0.2M & 2.4M* \\
    Gemma-4-31B-it~\cite{gemma4} & Adapt. & 2.3K & 0.6M$^\dagger$ \\
    Qwen3.6-27B~\cite{qwen36} & Adapt. & 65.5K & 16.8M \\
    \bottomrule
    \end{tabular}
\end{table}

To assess the correctness of the recognized characters and words, we employ Average Normalised Levenshtein Similarity (ANLS)~\cite{anls}, a standard quality proxy in OCR evaluation~\cite{docvqa, infovqa, ocrbench} that measures the character-level proximity between a model prediction and the reference transcription. We also use BLEU-1~\cite{bleu}, which rewards correct unigrams and is therefore resistant to different reading order interpretations from models.

\begin{table}[t]
    \centering
    \setlength{\tabcolsep}{5pt}
    \renewcommand{\arraystretch}{1.2}
    \caption{Performance on \benchname{} across subsets. We report ANLS ($\uparrow$) and BLEU-1 ($\uparrow$), both on a $0$--$100$ scale. R and FP denote the Region and Full Page Handwritten subsets, respectively. \textbf{Best} and \underline{second-best} per column are highlighted.}
    \label{tab:results_instruct}
    \resizebox{\textwidth}{!}{%
    \begin{tabular}{l@{\hspace{8pt}}cccccccccc}
    \toprule
    \multirow{2}{*}{\textbf{Model}} & \multicolumn{2}{c}{\textbf{Handwritten (R)}} & \multicolumn{2}{c}{\textbf{Handwritten (FP)}} & \multicolumn{2}{c}{\textbf{Synthetic}} & \multicolumn{2}{c}{\textbf{In-The-Wild}} & \multicolumn{2}{c}{\textbf{Mean}} \\
    \cmidrule(lr){2-3} \cmidrule(lr){4-5} \cmidrule(lr){6-7} \cmidrule(lr){8-9} \cmidrule(lr){10-11}
     & \textbf{ANLS} & \textbf{BLEU-1} & \textbf{ANLS} & \textbf{BLEU-1} & \textbf{ANLS} & \textbf{BLEU-1} & \textbf{ANLS} & \textbf{BLEU-1} & \textbf{ANLS} & \textbf{BLEU-1} \\
    \midrule
    \multicolumn{11}{@{}l}{\textbf{\textit{OCR-Specific}}} \\
    Nemotron-OCR-V2~\cite{nemotron_ocr} & 15.6 & 9.6 & 11.2 & 9.5 & 91.4 & 81.0 & 75.9 & 68.8 & 48.5 & 42.2 \\
    DeepSeek-OCR-2-3B~\cite{deepseek_ocr} & 36.1 & 30.2 & 18.7 & 23.3 & 91.4 & 84.3 & 84.1 & 75.0 & 57.6 & 53.2 \\
    \midrule
    \multicolumn{11}{@{}l}{\textbf{\textit{First-tier LVLMs ($\leq$ 12B parameters)}}} \\
    Molmo2-O-7B~\cite{molmo2} & 26.1 & 22.1 & 6.6 & 10.7 & 70.7 & 61.5 & 56.7 & 34.8 & 40.0 & 32.3 \\
    GLM-4.6V-Flash-10B~\cite{glm} & 46.5 & 34.3 & 21.8 & 19.3 & 56.8 & 54.1 & 48.3 & 45.5 & 43.4 & 38.3 \\
    PerceptionLM-8B~\cite{perceptionlm} & 36.7 & 29.9 & 10.1 & 9.1 & 78.2 & 71.7 & 59.2 & 46.0 & 46.1 & 39.2 \\
    Molmo2-8B~\cite{molmo2} & 35.0 & 26.8 & 12.4 & 14.3 & 81.5 & 66.6 & 70.9 & 44.0 & 50.0 & 37.9 \\
    AMALIA-VL~\cite{amalia-vl} & 41.6 & 32.7 & 28.4 & 28.0 & 81.3 & 79.6 & 72.6 & 60.8 & 56.0 & 50.3 \\
    InternVL3.5-8B~\cite{internvl} & 51.6 & 43.6 & 31.8 & 34.0 & 75.5 & 71.6 & 81.0 & 71.7 & 60.0 & 55.2 \\
    AMALIA-VL-DPO~\cite{amalia-vl} & 48.2 & 38.2 & 39.3 & 37.7 & 83.2 & 81.5 & 74.8 & 64.6 & 61.4 & 55.5 \\
    Bee-RL-8B~\cite{beerl} & 45.2 & 37.5 & 27.1 & 30.7 & 94.6 & 87.7 & 83.1 & 79.1 & 62.5 & 58.8 \\
    Nemotron-Nano-12B-VL-V2~\cite{nemotron_vl} & 53.3 & 42.3 & 37.5 & 32.7 & 92.2 & 90.0 & 76.8 & 72.9 & 65.0 & 59.5 \\
    LLaVA-OneVision-1.5-8B~\cite{llavaonevision15} & 61.1 & 51.4 & 32.6 & 37.2 & 98.2 & 93.0 & 77.4 & 70.9 & 67.3 & 63.1 \\
    Gemma-4-E4B-it~\cite{gemma4} & 78.2 & 63.7 & 42.7 & 38.7 & 95.6 & 89.1 & 82.7 & 78.5 & 74.8 & 67.5 \\
    Ministral3-Instruct-8B~\cite{ministral3} & 74.0 & 59.9 & 63.9 & 56.8 & 98.6 & 96.1 & 71.4 & 58.9 & 77.0 & 67.9 \\
    Qwen3.5-9B~\cite{qwen35} & 80.1 & 70.3 & \underline{70.0} & \underline{64.9} & \underline{99.8} & \textbf{99.0} & 85.4 & 80.9 & 83.8 & 78.8 \\
    \midrule
    \multicolumn{11}{@{}l}{\textbf{\textit{Second-tier LVLMs ($>$ 12B parameters)}}} \\
    InternVL3.5-38B~\cite{internvl} & 56.5 & 47.7 & 42.6 & 43.2 & 81.2 & 76.8 & 75.2 & 68.2 & 63.9 & 59.0 \\
    Gemma-4-31B-it~\cite{gemma4} & \textbf{88.0} & \textbf{79.0} & 65.2 & 58.8 & 98.3 & 93.0 & \underline{89.8} & \textbf{87.0} & \underline{85.3} & \underline{79.5} \\
    Qwen3.6-27B~\cite{qwen36} & \underline{81.5} & \underline{71.8} & \textbf{75.4} & \textbf{71.1} & \textbf{99.9} & \underline{96.5} & \textbf{90.5} & \underline{86.5} & \textbf{86.8} & \textbf{81.5} \\
    \bottomrule
    \end{tabular}%
    }
\end{table}

\begin{figure}[t]
    \centering
    \newlength{\panelheight}%
    \settoheight{\panelheight}{\includegraphics[width=0.47\columnwidth]{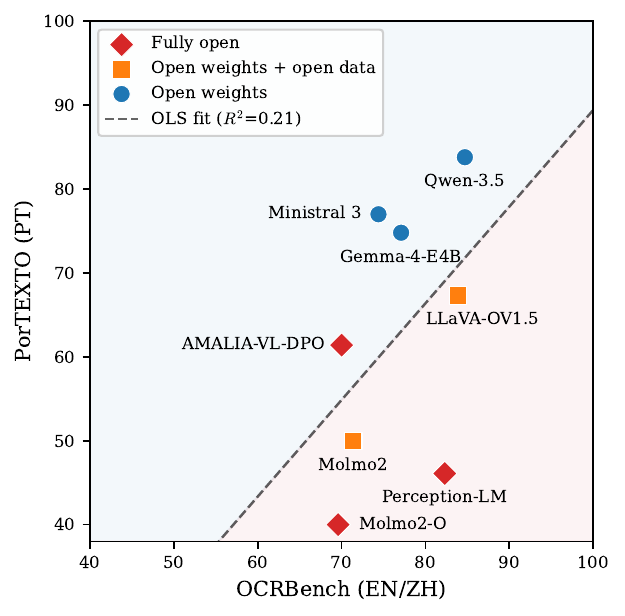}}%
    \begin{minipage}[t]{0.49\columnwidth}
        \centering
        \includegraphics[width=\linewidth]{figures/ocrbench_vs_portexto.pdf}
         \caption{OCRBench vs.\ \benchname{} mean ANLS; both axes on a $0$--$100$ scale. The dashed line is the OLS fit ($R^2 = 0.21$). 
        }
        \label{fig:ocrbench_corr}
    \end{minipage}\hfill
    \begin{minipage}[t]{0.47\columnwidth}
        \centering
        \begin{minipage}[b][\panelheight][b]{\linewidth}
            \centering
            \includegraphics[width=\linewidth]{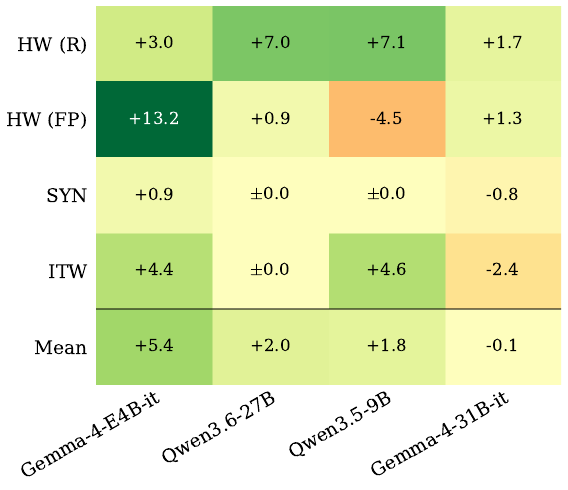}
        \end{minipage}
        \caption{Per-subset $\Delta$ANLS = (Reasoning $-$ Instruct). HW(R) is the Handwritten (Region) subset, HW (FP) is the Handwritten (Full Page) subset, SYN is the Synthetic subset and ITW is the In-The-Wild subset.
        }
        \label{fig:thinking_heatmap}
    \end{minipage}
\end{figure}
 
\section{Results and Discussion}
Table \ref{tab:results_instruct} shows the results of evaluating our selected model suite on \benchname, broken down by subset, metric, and model category.

Results show an overall \textbf{strong performance on the Synthetic subset, but a significant performance drop in In-The-Wild and Handwritten samples}. OCR models in particular collapse on the Handwritten subset. On In-The-Wild inputs, however, they remain competitive.

Qwen3.5-9B~\cite{qwen35} achieves the highest score in the first tier ($<=$ 12B). In the second tier ($>$12B), Qwen3.6-27B~\cite{qwen36} and Gemma-4-31B-it~\cite{gemma4} reach much superior scores than InternVL3.5-38B and other first tier models (except for Qwen3.5-9B). Notably, Ministral3-Instruct-8B~\cite{ministral3} and Gemma-4-E4B-it~\cite{gemma4} rank much higher than other models with significantly higher resolution budget (see Table \ref{tab:model_properties}) and even higher model sizes. 
Gemma-4 models have the lowest resolution budget\footnote{We used the model's default resolution budget.}, and yet reach the top-3 and top-2 scores in the small and large tier, respectively. 
Furthermore, the gap between the top performer in the small tier and the top performer in the large tier is small, hinting that model size and resolution budget are not the main bottleneck for pt-PT OCR performance. We do note, however, that within model family, larger models do achieve better performance.

\vspace{-2mm}\subsubsection{Evidencing pt-PT OCR model limitations.}
We hypothesize that specialized multilingual data is the main differentiator, since it is a common focus of all top performing models. The top performing open-source models in~\benchname~are open weights and do not reveal their training data mix. Open data models such as LLaVA-OneVision-1.5-8B~\cite{llavaonevision15}, Molmo2-8B~\cite{molmo2} and Bee-RL-8B~\cite{beerl} trail behind in performance. 
The two fully open models (i.e., fully open LLM and open vision data), Molmo2-O-7B~\cite{molmo2} and PerceptionLM-8B~\cite{perceptionlm}, are among the weakest models, achieving lower mean BLEU-1 than both OCR-specific models.
Thus, we further hypothesize that the lack of publicly available pt-PT centric OCR data is one of the main factors hurting these models. AMALIA-VL~\cite{amalia-vl} and AMALIA-VL-DPO~\cite{amalia-vl}, fully open models specializing in European Portuguese, perform best among fully open models and on par with open-data models, corroborating our hypothesis. To probe this further, we compute the performance of a subset of models in OCRBench~\cite{ocrbench}, a benchmark in the same spirit as \benchname, but only focused on two high-resource languages: English (EN) and Chinese (ZH). 
Figure \ref{fig:ocrbench_corr} shows a regression line predicting the model performance in pt-PT OCR given their performance in EN/ZH OCR. 
 
We observed no discernible relationship between EN/ZH and pt-PT scores ($R^2=0.21$). In fact, fully open and open data models performed competitively against open weights in EN/ZH, unlike in pt-PT. The most notable examples are PerceptionLM-8B and LLaVA-OneVision-1.5-8B, which surpassed the second and third best pt-PT models on EN/ZH and were competitive with the best performing model, but trailed significantly behind in pt-PT. AMALIA-VL-DPO, on the other hand, has poor EN/ZH performance but exceeds expectations for fully open models in pt-PT OCR. These results support our hypothesis, and motivate the need for future research on open pt-PT specific OCR training data.

\vspace{-2mm}\subsubsection{Impact of Reasoning in Visual Text Extraction.}
Figure~\ref{fig:thinking_heatmap} reports the performance delta induced by enabling explicit reasoning in unified models.
Three of the four models register a positive mean ANLS gain, and the largest mean improvement is attained by Gemma-4-E4B-it~\cite{gemma4}.
Gains on the Handwritten (Region) subset are positive for all four models, but the Handwritten (Full Page) and In-The-Wild subsets show mixed results. Overall, reasoning appears to be a useful lever for improving pt-PT visual text extraction, but the performance decrease in some models for the most challenging subsets may also hint at the possibility that they can struggle to think with challenging visual pt-PT text. Future research is needed to better understand this phenomenon.

\section{Conclusion}

We introduced \benchname, a novel benchmark that focuses on contemporary and culturally relevant pt-PT text extraction from images. 
Our benchmark comprehensively covers multiple OCR sources, including handwritten notes, in-the-wild photographs, and text overlays across a wide range of backgrounds, spanning pt-PT casual and formal use cases. Data quality is ensured by a carefully designed annotation pipeline, with manual review of native pt-PT speakers across all steps.
We established the first systematic assessment of visual text extraction performance on modern pt-PT text, exposing the difficulties of modern LVLMs to extract pt-PT text in real world inputs. \benchname~ fills this gap, and exposes the fragilities of open weights and open data LVLMs on OCR extraction in pt-PT.
We believe that the challenges signaled by \benchname, coupled with the rising demand for pt-PT visual text extraction, will encourage future research in the field.

\section{Acknowledgements}
This work was supported by the AMALIA project under Measure RE-C05-i08 of the Portuguese national ``Programa de Recuperação e Resiliência''. We also acknowledge the support of Fundação para a Ciência e Tecnologia (FCT) and the NOVA LINCS project (UID/04516/2025). We thank the Barcelona Supercomputing Center (BSC) for providing the computational resources that made this work possible.

\bibliographystyle{splncs04}
\bibliography{bibliography}

@article{amalia-vl,
  title={AMALIA-VL: A Native European Portuguese Open-Source Vision and Language Model},
  author={Gl{\'o}ria-Silva, Diogo and others},
  journal={arXiv:2606.19100},
  year={2026}
}

@article{ocrbench,
  title={Ocrbench: on the hidden mystery of ocr in large multimodal models},
  author={Liu, Yuliang and others},
  journal={Science China Information Sciences},
  year={2024},
  publisher={Springer}
}

@inproceedings{docvqa,
  author    = {Mathew, Minesh and others},
  title     = {{DocVQA}: A Dataset for {VQA} on Document Images},
  booktitle = {WACV},
  year      = {2021}
}

@inproceedings{infovqa,
  author    = {Mathew, Minesh and others},
  title     = {{InfographicVQA}},
  booktitle = {WACV},
  year      = {2022}
}

@inproceedings{omnidocbench,
  author    = {Ouyang, Linke and others},
  title     = {{OmniDocBench}: Benchmarking Diverse {PDF} Document Parsing with Comprehensive Annotations},
  booktitle = {CVPR},
  year      = {2025}
}

@inproceedings{mtvqa,
  author    = {Tang, Jingqun and others},
  title     = {{MTVQA}: Benchmarking Multilingual Text-Centric Visual Question Answering},
  booktitle = {ACL Findings},
  year      = {2025}
}

@inproceedings{cc-ocr,
  author    = {Yang, Zhibo and others},
  title     = {{CC-OCR}: A Comprehensive and Challenging {OCR} Benchmark for Evaluating Large Multimodal Models in Literacy},
  booktitle = {ICCV},
  year      = {2025}
}

@inproceedings{bressay,
  author    = {Neto, Arthur F. S. and others},
  title     = {{BRESSAY}: A {Brazilian Portuguese} Dataset for Offline Handwritten Text Recognition},
  booktitle = {ICDAR},
  year      = {2024}
}

@inproceedings{ester,
  author    = {Santos, Moniele Kunrath and others},
  title     = {{ESTER-Pt}: An Evaluation Suite for {TExt} Recognition in {Portuguese}},
  booktitle = {ICDAR},
  year      = {2023}
}

@inproceedings{porto,
  author    = {Os{\'o}rio, Tom{\'a}s Freitas and others},
  title     = {{Portuguese} post-{OCR} Resources for Text Optimisation},
  booktitle = {CIKM},
  year      = {2025}
}

@article{iforal,
  author  = {Matos, Alexandre and others},
  title   = {i{Foral}: Automated Handwritten Text Transcription for Historical Medieval Manuscripts},
  journal = {Journal of Imaging},
  year    = {2025}
}

@misc{mazaafard,
  author       = {{mazafard}},
  title        = {Portuguese {OCR} Dataset},
  year         = {2025},
  howpublished = {Hugging Face, \url{https://huggingface.co/datasets/mazafard/portuguese-ocr-dataset}}
}

@misc{gemini,
  author       = {{Google DeepMind}},
  title        = {Gemini 3.1 Pro Model Card},
  year         = {2026},
  howpublished = {\url{https://deepmind.google/models/model-cards/gemini-3-1-pro/}},
}

@inproceedings{textvqa,
  author    = {Singh, Amanpreet and others},
  title     = {Towards {VQA} Models that can Read},
  booktitle = {CVPR},
  year      = {2019}
}

@inproceedings{okvqa,
  author    = {Marino, Kenneth and others},
  title     = {{OK-VQA}: A Visual Question Answering Benchmark Requiring External Knowledge},
  booktitle = {CVPR},
  year      = {2019}
}

@misc{nemotron_ocr,
  author       = {Liu, Bo and others},
  title        = {Building a Fast Multilingual {OCR} Model with Synthetic Data},
  year         = {2026},
  howpublished = {NVIDIA, \url{https://huggingface.co/blog/nvidia/nemotron-ocr-v2}},
}

@misc{episa,
  author       = {EPISA Project},
  title        = {Typewritten Digital Representations of {Portuguese} Cultural Heritage Documents from the 20th Century ({EPISA} Dataset)},
  howpublished = {INESC TEC, \url{https://episa.inesctec.pt/outcomes/}},
  year         = {2022}
}

@misc{wikimedia2023wikipedia,
    title        = {Wikimedia Wikipedia (20231101.pt)},
    author       = {{Wikimedia Foundation}},
    year         = {2023},
    howpublished = {HuggingFace Datasets},
    url          = {https://huggingface.co/datasets/wikimedia/wikipedia}
}

@inproceedings{koehn2005europarl,
    title     = {Europarl: A Parallel Corpus for Statistical Machine Translation},
    author    = {Koehn, Philipp},
    booktitle = {MT},
    year      = {2005},
    url       = {https://www.statmt.org/europarl/}
  }

@inproceedings{synthdog,
  author    = {Kim, Geewook and others},
  title     = {{OCR}-free Document Understanding Transformer},
  booktitle = {ECCV},
  year      = {2022},
}

@article{openimages,
  author  = {Kuznetsova, Alina and others},
  title   = {The {Open Images Dataset V4}: Unified Image Classification, Object Detection, and Visual Relationship Detection at Scale},
  journal = {IJCV},
  year    = {2020}
}

@inproceedings{lmms_eval,
  title={Lmms-eval: Reality check on the evaluation of large multimodal models},
  author={Zhang, Kaichen and others},
  booktitle={NAACL},
  year={2025}
}

@inproceedings{anls,
  author    = {Biten, Ali Furkan and others},
  title     = {Scene Text Visual Question Answering},
  booktitle = {ICCV},
  year      = {2019}
}

@inproceedings{bleu,
  author    = {Papineni, Kishore and others},
  title     = {{BLEU}: a Method for Automatic Evaluation of Machine Translation},
  booktitle = {ACL},
  year      = {2002}
}

@misc{qwen35,
  author       = {{Qwen Team}},
  title        = {{Qwen3.5}: Towards Native Multimodal Agents},
  year         = {2026},
  howpublished = {\url{https://qwen.ai/blog?id=qwen3.5}}
}

@misc{qwen36,
  author       = {{Qwen Team}},
  title        = {{Qwen3.6-27B}: Flagship-Level Coding in a {27B} Dense Model},
  year         = {2026},
  howpublished = {\url{https://qwen.ai/blog?id=qwen3.6-27b}},
}

@misc{gemma4,
howpublished = {\url{https://ai.google.dev/gemma/docs/core/model\_card\_4}},
  author       = {{Google DeepMind}},
  title        = {{Gemma 4} Model Card},
  year         = {2026}
  }

@article{llavaonevision15,
  author        = {An, Xiang and others},
  title         = {{LLaVA-OneVision-1.5}: Fully Open Framework for Democratized Multimodal Training},
  year          = {2025},
  journal       = {arXiv:2509.23661}
}

@article{perceptionlm,
  title={Perceptionlm: Open-access data and models for detailed visual understanding},
  author={Cho, Jang Hyun and others},
  journal={Advances in Neural Information Processing Systems},
  year={2026}
}

@article{molmo2,
  title={Molmo2: Open Weights and Data for Vision-Language Models with Video Understanding and Grounding},
  author={Clark, Christopher and others},
  journal={arXiv:2601.10611},
  year={2026}
}

@article{beerl,
  author        = {Zhang, Yi and others},
  title         = {{Bee}: A High-Quality Corpus and Full-Stack Suite to Unlock Advanced Fully Open {MLLMs}},
  year          = {2025},
  journal={arXiv:2510.13795},
}

@article{ministral3,
  title={Ministral 3},
  author={Liu, Alexander H and others},
  journal={arXiv:2601.08584},
  year={2026}
}

@inproceedings{euroeval,
  title     = {Encoder vs decoder: Comparative analysis of encoder and decoder language models on multilingual {NLU} tasks},
  author    = {Nielsen, Dan Saattrup and others},
  booktitle = {NoDaLiDa/Baltic-HLT},
  year      = {2025}
}

@article{deepseek_ocr,
  title={DeepSeek-OCR 2: Visual Causal Flow},
  author={Wei, Haoran and others},
  journal={arXiv:2601.20552},
  year={2026}
}

@article{nemotron_vl,
  title={Nvidia nemotron nano v2 vl},
  author={Deshmukh, Amala Sanjay and others},
  journal={arXiv:2511.03929},
  year={2025}
}

@article{glm,
  title={Glm-4.5 v and glm-4.1 v-thinking: Towards versatile multimodal reasoning with scalable reinforcement learning},
  author={Hong, Wenyi and others},
  journal={arXiv:2507.01006},
  year={2025}
}

@article{internvl,
  title={Internvl3. 5: Advancing open-source multimodal models in versatility, reasoning, and efficiency},
  author={Wang, Weiyun and others},
  journal={arXiv:2508.18265},
  year={2025}
}

\end{document}